\documentclass[11pt]{article}

\usepackage[preprint]{acl}

\usepackage{times}
\usepackage{latexsym}
\usepackage[T1]{fontenc}
\usepackage[utf8]{inputenc}
\usepackage{microtype}
\usepackage{inconsolata}
\usepackage{graphicx}
\usepackage{amsmath}
\usepackage{amssymb}
\usepackage{booktabs}
\usepackage{multirow}
\usepackage{array}
\usepackage{tabularx}
\usepackage{placeins}

\let\hermesincludegraphics\includegraphics
\renewcommand{\includegraphics}[2][]{%
  \IfFileExists{#2}{%
    \hermesincludegraphics[#1]{#2}%
  }{%
    \fbox{\parbox{0.9\linewidth}{\centering Missing figure: \texttt{\detokenize{#2}}}}%
  }%
}

\title{HERMES: A Multi-Granularity Labeling Substrate for Pre-training Data Mixtures}

\author{
\begin{minipage}{0.98\textwidth}
\centering
Ziyun Qiao\textsuperscript{1,2,*,\dag}
\quad
Yue Min\textsuperscript{1,*}
\quad
Ruining Chen\textsuperscript{3}
\quad
Yujun Li\textsuperscript{1,\dag}
\\[0.35em]
\normalfont\textsuperscript{1}\,Wizard Quant, Beijing, China
\quad
\textsuperscript{2}\,Peking University, Beijing, China
\quad
\textsuperscript{3}\,University of Science and Technology of China, Hefei, China
\\[0.25em]
\normalfont\textsuperscript{*}Equal contribution.
\\[0.15em]
\normalfont\textsuperscript{\dag}Correspondence to:
\texttt{qiaozy@stu.pku.edu.cn}, \texttt{liyujun@wizardquant.com}
\end{minipage}
}

\begin{document}
\maketitle

\begin{abstract}
Most data-mixing methods assume the corpus has already been partitioned into groups, and the choice of those groups determines what a mixer can express. Existing labels (provenance, topic or format taxonomies, flat embedding clusters) commit to one semantic axis at one granularity; changing the resolution rebuilds the labels. We argue the bottleneck is the label system, not the mixer, and provide a hierarchical one. \textbf{HERMES} is a data-derived labeling substrate: a Learned Semantic Transform followed by 3-stage residual vector quantization annotates each document once into a coarse-to-fine code whose prefix length controls granularity up to ${\sim}130$k cells. At coarse granularity HERMES sits at a plateau with KMeans-family methods on standard clustering metrics, so the contribution is the substrate, not the clusterer. On 1B-parameter, 25B-token pre-training, the hierarchy exposes an interaction fixed-granularity pipelines cannot test: at one prefix length, a combined Stage-2 rule contrast (equal-subbucket coverage vs.\ size-proportional within-bucket quality top-30\%) lifts a 16-task capability macro-average by $+0.0253$; at the next finer level, the same rule loses its measurable edge as candidate pools contract ${\sim}5\times$. HERMES reframes data mixture design from choosing among fixed label sets to navigating a reusable, data-derived granularity hierarchy.
\end{abstract}

\section{Introduction}
\label{sec:intro}

\begin{figure}[t]
\centering
\includegraphics[width=\columnwidth]{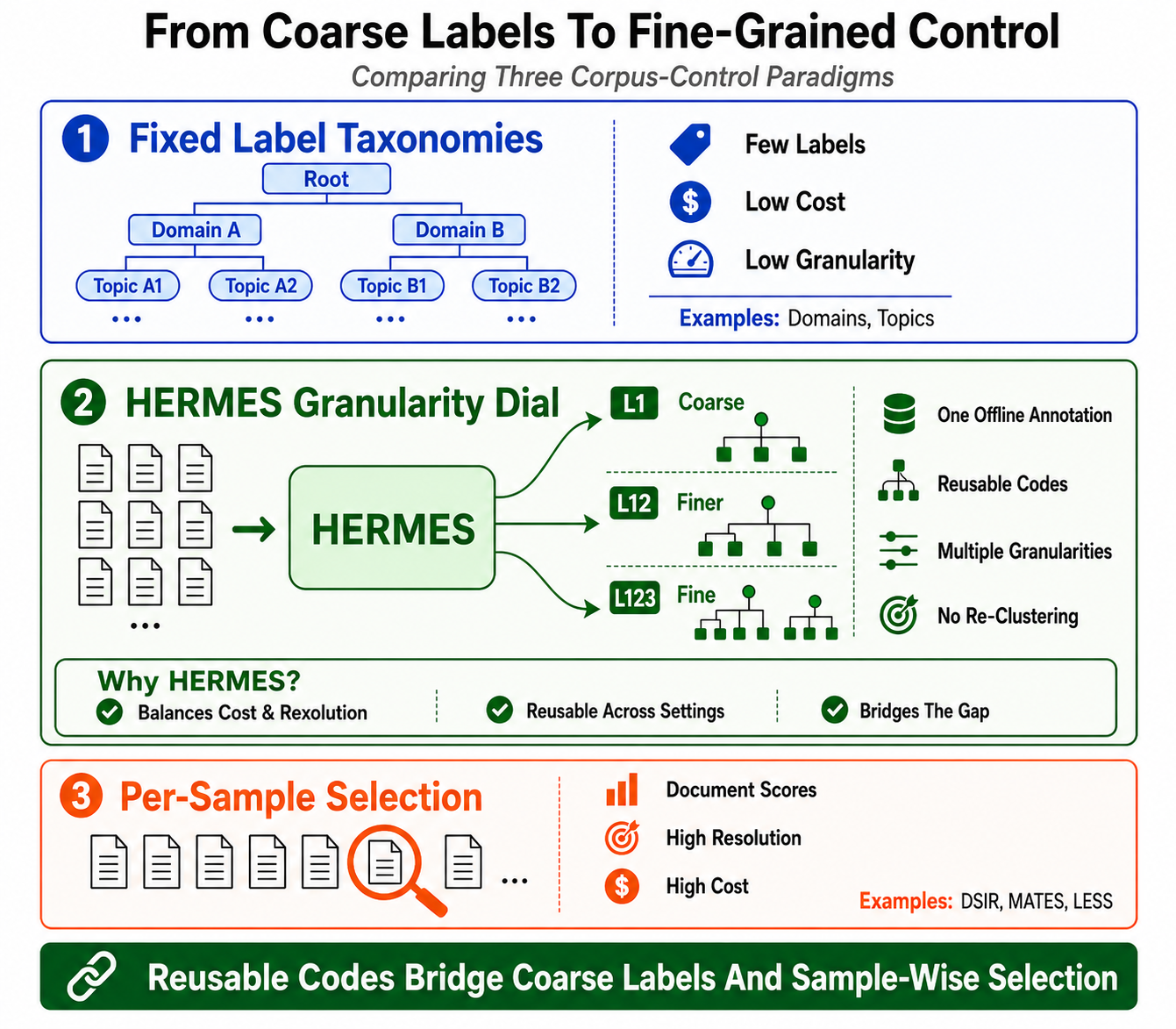}
\caption{Three corpus-control paradigms. \emph{Top:} fixed taxonomies (source, topic, format) scale but fix one semantic axis at one granularity. \emph{Bottom:} per-sample selection (DSIR, MATES, LESS) reaches the document at per-document compute. \emph{Middle:} HERMES exposes a data-derived hierarchy whose prefix reads ($L_1, L_{12}, L_{123}$) deliver multiple granularities from one offline annotation, with no re-clustering.}
\label{fig:cost-resolution}
\end{figure}

Data curation has become a decisive factor in LLM pre-training \citep{refinedweb,fineweb-edu}, shifting attention from parameter scaling alone toward the mixing of heterogeneous sources. Pre-training data mixing has two separable layers: a \emph{label system} that partitions the corpus, and a \emph{mixer or sampler} that consumes those labels. What has constrained group-level pipelines is the label layer, not the mixer machinery: provenance is coarse, taxonomies commit to a single semantic axis and granularity, and flat clusterings require recomputation to change $K$. We address this with \textbf{HERMES} (Hierarchical Embedding-based RVQ for Mixture Selection), a one-shot hierarchical residual vector quantizer that annotates each document with a coarse-to-fine code $(c_1,c_2,c_3)$ and exposes granularity as a prefix length on a single trained RVQ codebook stack (Figure~\ref{fig:cost-resolution}). The detailed comparison to existing provenance, distilled-taxonomy, and embedding-cluster label families is deferred to Section~\ref{sec:related}.

At the $K=256$--$130$k label scale, regression-style mixture optimizers (RegMix \citep{regmix}, Data Mixing Laws \citep{datamixinglaws}) sit outside the proxy budget, since their fitting bases scale at least linearly in cell count; we therefore instantiate only $O(1)$-proxy outer weights, \emph{Uniform} and \emph{DoReMi}. Per-sample selectors that avoid labels pay per-document compute at web scale (Section~\ref{sec:baselines}).

HERMES meets these constraints. A Learned Semantic Transform, in the spirit of optimized product quantization \citep{opq}, feeds a 3-stage residual vector quantizer \citep{irvq,qinco}; the prefix length on the resulting code defines granularity over a single trained RVQ codebook stack: $L_1=c_1$ gives 256 buckets, $L_{12}$ gives ${\sim}65$k, $L_{123}$ gives ${\sim}130$k observed. HERMES does not aim to be a better clustering algorithm: at $K=256$ it sits on a plateau with four standard 256-way clusterers on compactness/mass-balance metrics (Table~\ref{tab:plateau}), and is downstream-matched to KMeans within $0.0002$ Avg.\ (Section~\ref{sec:plateau}). Its value is that one annotation supports an entire granularity sweep with no re-clustering between $K=256$ and $K=130$k.

HERMES exposes two interactions invisible to fixed-granularity pipelines, on a 1B/25B-token training regime evaluated by a 16-task capability macro-mean (Avg.). First, at fixed $L_{12}$ granularity under DoReMi-L1, switching the Stage-2 sampler from max-entropy coverage (equal sub-bucket mass, all docs) to quality top-30\% (size-proportional mass, within-bucket top-30\%; corrected FineWeb-Edu reader) raises Avg.\ by $+0.0253$ (z $+2.09$) at fixed corpus, model, codebook, and outer weight (Section~\ref{sec:findingA}). Second, the same Stage-2 advantage does not persist at $L_{123}$: under the quota-preserving pair (per-L1 $L_{123}$ local random coverage vs.\ $L_{123}$ quality top-30\%), the two rules sit at $0.3986$ and $0.3988$ ($\Delta=+0.0002$, numerically tied under single-seed evaluation; Section~\ref{sec:mechanism}). This collapse is consistent with candidate competition: the median candidate pool contracts $5.3\times$ from $L_{12}$ (2{,}271) to $L_{123}$ (429), a regime in which within-sub-bucket ranking is no longer a stable proxy for global ranking over coverage. Together these show that granularity and the Stage-2 sampler are jointly determined on the substrate; a paired Uniform vs.\ DoReMi $L_1$ characterisation is in Appendix~\ref{app:outerweight}.

\textbf{Contributions.} (i) HERMES, a one-shot hierarchical RVQ substrate that exposes granularity as a prefix length on a trained codebook stack, annotating ${\sim}50$M documents into up to ${\sim}130$k cells. (ii) A controlled Stage-2 sampler study at fixed codebook, corpus, model, budget, and granularity ($L_{12}$ under DoReMi-L1): switching from max-entropy coverage to quality top-30\% (corrected reader) yields $+0.0253$ Avg.\ (Finding A); the contrast moves two axes jointly (sub-bucket mass and per-document eligibility) and is read as a combined Stage-2 rule effect. (iii) A candidate-competition diagnostic: holding DoReMi-L1 outer weights fixed and comparing the same two rules in their quota-preserving forms, the L12 advantage ($+0.0253$) collapses to $+0.0002$ at $L_{123}$ ($0.3988$ vs.\ $0.3986$ per-L1 local random coverage), consistent with median pool shrinkage from $2{,}271$ to $429$ documents (Finding B, Section~\ref{sec:mechanism}).

\begin{figure*}[t]
\centering
\includegraphics[width=0.98\textwidth]{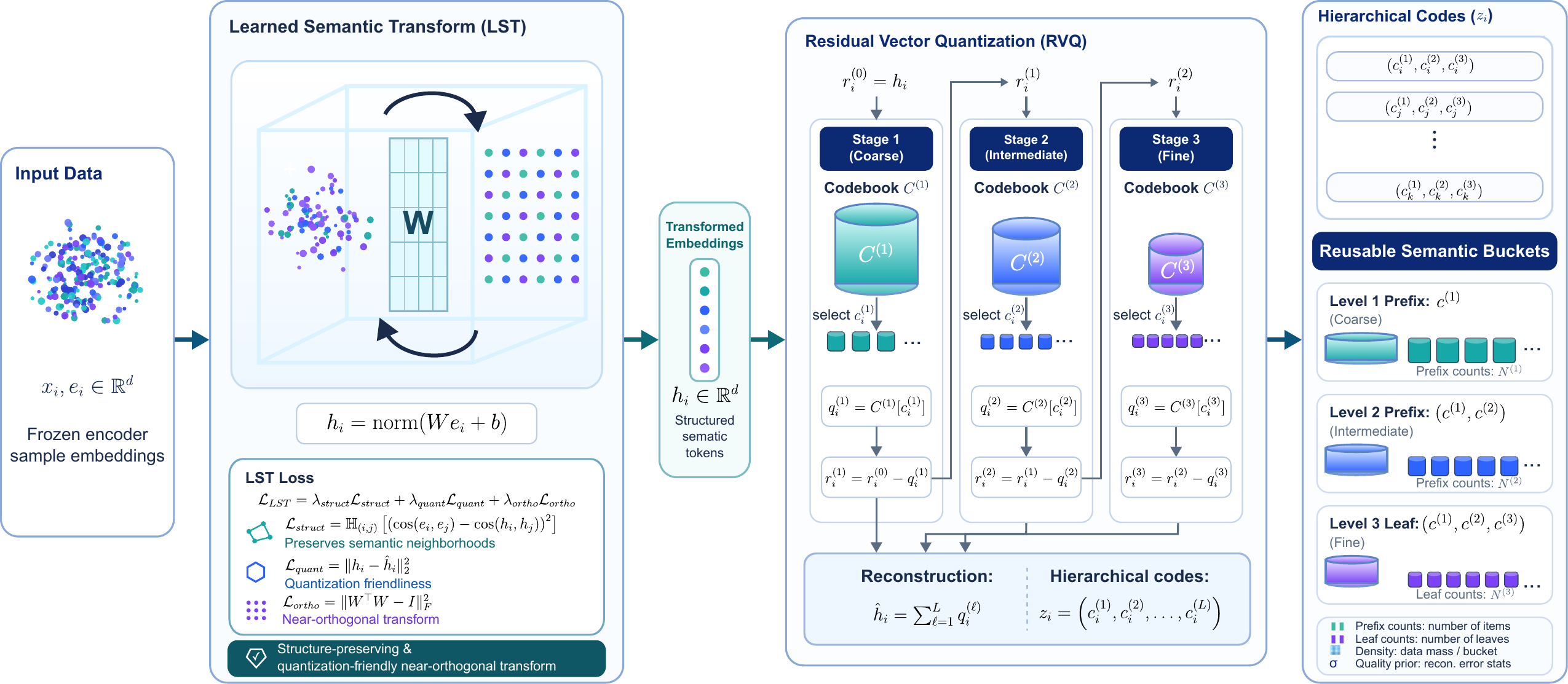}
\caption{HERMES annotation pipeline. A frozen encoder produces document embeddings (\emph{left}); the Learned Semantic Transform rotates them into a quantization-friendly geometry under structure-preservation, reconstruction, and orthogonality losses (\emph{centre-left}). Three cascaded RVQ stages then encode each rotated embedding into a hierarchical code (\emph{centre-right}). Reading the code at different prefix lengths (\emph{right}) exposes ${\sim}256$, ${\sim}65$k, and ${\sim}130$k buckets, reused across samplers without re-clustering. Sections~\ref{sec:lst}--\ref{sec:rvq}.}
\label{fig:teaser}
\end{figure*}

\section{Related Work}
\label{sec:related}

\textbf{Corpus label systems.} The label system used to partition a corpus is a design choice independent of the mixer that consumes it. Prior work falls into three families. \emph{Provenance} labels use the source domain or web shard a document originated from \citep{doremi}, easy to obtain but committed to a single coarse axis. \emph{Distilled taxonomies} build human- or LLM-defined category sets at a fixed granularity: WebOrganizer \citep{weborganizer} releases topic and format taxonomies trained on LLM annotations, and Topic-over-Source \citep{topicoversource} derives topic labels via clustering, LLM summarization, and classifier distillation, at a single fixed topical resolution. \emph{Embedding clusters} are unsupervised and data-derived: KMeans variants, plain RVQ \citep{irvq}, and CLIMB's ${\sim}20$ flat cells \citep{climb}. These three families differ along prior structure, semantics, and granularity. Topic-over-Source shares the data-derived prior but yields a single fixed topic level via an LLM-in-the-loop pipeline; HERMES exposes granularity as a prefix length on one quantizer pass, with no LLM in the loop.

\textbf{Mixers and samplers that consume labels.} Given a label system, several outer-weight families decide how much sampling probability each cell receives. DoReMi \citep{doremi} runs group-DRO on a small reference model. Data Mixing Laws \citep{datamixinglaws} and RegMix \citep{regmix} fit regression proxies whose bases scale linearly in cell count and are typically reported with ${\le}20$ groups. CLIMB \citep{climb} couples clustering and weight search in a joint loop. Chameleon \citep{chameleon} computes leverage-score weights over a learned domain-affinity matrix. None of these works treat the label system as a separately tunable axis: label resolution is a property of the labeling step they presuppose, not a dial they sweep. HERMES is complementary: it provides the label substrate, and any of these outer-weight families can in principle consume it at a chosen prefix length, subject to their proxy budget. In this paper we plug in two $O(1)$-proxy families compatible with $K=256$--$130$k cells: Uniform and DoReMi. Per-sample selectors instead score documents individually (Section~\ref{sec:baselines}).

\textbf{Vector quantization machinery.} HERMES borrows the rotate-then-quantize machinery of optimized product quantization \citep{opq}, improved RVQ \citep{irvq}, and VQ-VAE \citep{vqvae}, all of which optimize \emph{reconstruction}. We reuse it for labeling: the code is consumed by samplers rather than a decoder, judged by the data-mixing interactions it exposes; see \citet{data-mixing-survey-2026} for the broader static-vs-dynamic landscape.

\section{Method}
\label{sec:method}

\subsection{Notation and overview}
\label{sec:notation}
HERMES is designed for the middle of the cost-resolution spectrum: one annotation per document, computed offline against frozen embeddings, that supports a prefix-length granularity dial up to ${\sim}130$k cells without re-clustering. Let $\mathcal{D}=\{x_i\}_{i=1}^N$ with each document mapped to a fixed embedding $e_i\in\mathbb{R}^d$ by a frozen encoder ($d=1024$, $N\approx 5\times 10^7$); embeddings are computed once and never updated. HERMES is a two-stage offline annotator with $K=256$ throughout,
\[
e_i \;\xrightarrow{\;\mathrm{LST}\;}\; h_i \in \mathbb{R}^d \;\xrightarrow{\;\mathrm{RVQ}\;}\; (c_1, c_2, c_3) \in [K]^3,
\]
a Learned Semantic Transform (LST) rotates the embedding into a quantization-friendly $h_i$ and a 3-stage Residual Vector Quantizer (RVQ) emits a hierarchical code. We write the bucket id at prefix length $\ell$ as $b_\ell(x_i) = (c_1, \ldots, c_\ell)$, so $b_1$ is an ancestor of $b_2$ which is an ancestor of $b_3$. We refer to the three prefix granularities as \textbf{L1}, \textbf{L12}, and \textbf{L123} and use prefix length as the granularity dial. Codes are produced \emph{once}, so any difference across samplers cannot be attributed to a different grouping.

\subsection{Learned Semantic Transform (LST)}
\label{sec:lst}
LST is a single linear-plus-normalize layer $h_i = \mathrm{normalize}(W e_i + b)$ with $W\in\mathbb{R}^{d\times d}$ initialized to the identity. It is trained jointly with the RVQ codebooks under three named losses (pairwise structure preservation, quantization-aware reconstruction, and orthogonality), following the rotate-then-quantize lineage of \citet{opq}; full equations and the per-stage commitment loss are in Appendix~\ref{app:hermes-train}. Optimizer and schedule are also in Appendix~\ref{app:hermes-train}.

\subsection{Residual Vector Quantization}
\label{sec:rvq}
Residual vector quantization (RVQ) is a workhorse for hierarchical discrete representations across audio coding \citep{soundstream,encodec}, approximate nearest-neighbour search \citep{pq,opq,irvq}, and generative modelling \citep{vqvae,rqvae}; we repurpose it here as a grouping substrate for pre-training data. The transformed vector $h_i$ is encoded by $L$ cascaded vector quantizers, each with codebook $e^{(k)}\in\mathbb{R}^{K\times d}$. Writing $r_1 = h_i$, stage $k$ picks $c_k = \arg\max_{j\in[K]}\cos(r_k, e^{(k)}_j)$, emits $q_k = e^{(k)}_{c_k}$, and passes residual $r_{k+1} = r_k - q_k$ to the next stage. The reconstruction $\hat h_i = \sum_{k=1}^{L} q_k$ enters only $\mathcal{L}_\text{quant}$; the downstream label is the code $(c_1,\ldots,c_L)$, and prefix lengths $\ell < L$ name strictly coarser partitions. Codebooks are updated by EMA with $k$-means initialization and a per-stage stop-gradient commitment term (full form in Appendix~\ref{app:hermes-train}). We set $L=3$ and $K=256$ in the main experiments; the L1 capacity ablation (Appendix~\ref{app:capacity}) varies $K$. After training, encoding is a deterministic argmax against frozen codebooks. Hierarchy depth $L$ and per-stage codebook size $K$ are themselves design axes; their useful settings are corpus-bounded by candidate competition (Section~\ref{sec:mechanism}), and we discuss this in Limitations.

\begin{table}[t]
\centering
\caption{Three granularities defined by code prefix. L123 is naturally sparse: only ${\sim}0.77\%$ of nominal triples are populated, as the stage-3 residuals concentrate on a thin manifold of the joint code space (Appendix~\ref{app:granularities}).}
\label{tab:granularities}
\small
\begin{tabularx}{\columnwidth}{Xlrr}
\toprule
Granularity & Code & Nominal & Observed \\
\midrule
L1 & $c_1$ & 256 & $256$ \\
L12 & $(c_1,c_2)$ & 65{,}536 & 65{,}408 \\
L123 & $(c_1,c_2,c_3)$ & $1.677\!\times\!10^{7}$ & 129{,}955 \\
\bottomrule
\end{tabularx}
\end{table}

\subsection{The sampling taxonomy}
\label{sec:sampling}
A sampler is the composition of two stages. For a target granularity level $\ell$ and a document $x$ with prefix codes $b_{\leq \ell}(x)=(b_1,\ldots,b_\ell)$, the sampling probability factorises as
\[
P\!\left(x \mid \mathrm{sampler}\right) \;=\; w_{b_1} \,\cdot\, \pi_{b_1}\!\bigl(b_{\leq \ell}\bigr) \,\cdot\, \rho_{b_{\leq \ell}}(x),
\]
where $w_{b_1}$ is the Stage-1 outer weight on the L1 ancestor, $\pi_{b_1}(\cdot)$ a Stage-2 distribution over that ancestor's active level-$\ell$ descendants, and $\rho_B(x)$ a per-document rule inside leaf sub-bucket $B$. We separate the two stages so the trade-offs between outer-weight family ($w$) and per-granularity sub-bucket rule ($\pi,\rho$) are unambiguous.

\paragraph{Stage 1: L1 outer weight.} Each of the 256 L1 buckets receives a non-negative weight summing to 1.
\emph{Uniform}: equal-per-bucket, the max-entropy choice on L1 and the canonical no-outer-learning baseline.
\emph{DoReMi}: weights from group-DRO at $g=L_1$ on a 120M-parameter / 2.5B-token proxy \citep{doremi,gdro}, learned once at L1 and reused without further learning in every non-Uniform HERMES experiment. WebOrganizer rows under DoReMi use group-DRO computed natively on the WO Topic/Format labels. These are the only two $O(1)$-proxy outer-weight families tractable at $K=256$ (Section~\ref{sec:intro}). We evaluate both at $L_1$ as a sanity audit; the paired characterisation is reported in Appendix~\ref{app:outerweight}. Our main granularity claim is conditioned on \emph{fixed} DoReMi outer weights, not on DoReMi being optimal.

\paragraph{Stage 2: Per-granularity sub-bucket sampler.} Stage 2 decides which sub-buckets inside each L1 ancestor are eligible and how documents are drawn inside them; it does \emph{not} re-distribute Stage-1 L1 weights. Samplers differ along two axes: sub-bucket mass is \emph{size-proportional} or \emph{max-entropy} (equal regardless of size); per-document eligibility is \emph{all} or \emph{top-30\% by FineWeb-Edu quality score} within that sub-bucket. Concretely, letting $\mathcal{C}_\ell(b_1)$ denote the L1 ancestor's active level-$\ell$ children, \emph{max-entropy} samplers set
\[
\pi_{b_1}(B) = \tfrac{1}{|\mathcal{C}_\ell(b_1)|}, \qquad \rho_B(x) = \tfrac{1}{|B|},
\]
and \emph{quality top-30\%} samplers set
\[
\pi_{b_1}(B) \propto |B|, \qquad \rho_B(x) = \tfrac{\mathbf{1}\{q(x)\geq\tau_B\}}{Z_B},
\]
where $q(\cdot)$ is the FineWeb-Edu quality score, $\tau_B$ is the within-sub-bucket 70th-percentile, and $Z_B$ normalises within $B$. We use four main Stage-2 samplers on the HERMES substrate: \emph{L1 max-entropy} (no sub-bucket structure below L1); \emph{L12 max-entropy} (within each L1 ancestor, equal-weight its $L_{12}$ children); \emph{L12 quality top-30\%} (corrected reader); and \emph{L123 quality top-30\%} (corrected reader). Diagnostic variants (\emph{L1-local quality top-30\%} for the granularity arc) and $L_{123}$ side-references (\emph{per-L1 $L_{123}$ local random coverage}, the quota-preserving coverage row for Finding B, and quota-flattening \emph{global $L_{123}$ max-entropy}) are defined in Appendix~\ref{app:sampling-variants}; they appear in the leaderboard (Appendix~\ref{app:leaderboard}).

A configuration is (L1 outer family) $\times$ (Stage-2 sampler). All controlled experiments share the same HERMES-$256^3$ codebook. The within-sub-bucket quality samplers read the FineWeb-Edu quality score via a per-key bounded heap; a corrected-reader specification and a 70th-percentile sanity number are in Appendix~\ref{app:quality}. All quality top-30\% rows we discuss in Findings A and B use the corrected reader.

\section{Experimental Setup}
\label{sec:setup}

\subsection{Pre-training}
\label{sec:pretrain}
We train a 1B-parameter LLaMA-style decoder \citep{llama} for 25B tokens on each sampler configuration. All runs share the same architecture, optimizer, learning-rate schedule, context length, tokenizer, and token budget; only the upstream sampler varies. The pre-training corpus is an internal ${\sim}50$M-document collection pre-filtered with public quality classifiers; we treat it as a fixed source distribution and study how to sample from it, not how to clean it. Sampling is with replacement, so the realised sampled set can exceed the source corpus (see Appendix~\ref{app:sampling}). Embeddings used by HERMES are produced once by the same frozen encoder for every document and are never updated. We report 20 ranked 1B/25B checkpoints (14 main rows + 6 HERMES capacity-ablation rows) in Appendix~\ref{app:leaderboard}, spanning granularity, Stage-2 sampler, L1-outer-weight, and grouping-method axes; the full leaderboard is sorted by Avg.

\subsection{Downstream evaluation}
\label{sec:eval}
We evaluate on 16 capability sub-tasks listed in Appendix~\ref{app:olmes}, grouped into four ability families: Basic Skills (6 probe sub-tasks, from \citealp{olmes}), Science QA (5; ARC, SciQ, PIQA, LAB-Bench \citep{arc,sciq,piqa,labbench}), Language Modeling (HellaSwag \citep{hellaswag}), and Others (CommonsenseQA \citep{csqa}, Jeopardy, NaturalQuestions \citep{nq}, Social IQA \citep{siqa}). Our headline number is \textbf{Avg.}, the family-equal macro-mean accuracy
\[\mathrm{Avg.}=\tfrac{1}{4}\sum_{g}\overline{\mathrm{acc}_g},\]
which prevents the six Basic Skills sub-tasks from dominating. A secondary \textbf{z-score} column standardises each sub-task across the 20 ranked checkpoints and averages within each family. The z column is a unit conversion for legibility, not a separate metric; Avg.\ and z are reported together throughout.

\subsection{Baselines}
\label{sec:baselines}
\emph{Learned groupings at $L_1{=}256$.} KMeans-256, MiniBatchKMeans-256, BisectingKMeans-256 (all from scikit-learn \citep{sklearn,lloyd-kmeans,minibatch-kmeans,bisecting-kmeans}), Plain RVQ-$256^3$ (RVQ without LST), and HERMES-$256^3$ (ours). All non-HERMES learned groupings share an identical annotation pipeline (fit on a 1.16M-document subsample at $K=256$ with cosine assignment, then deterministic full-corpus labeling) so that only the clustering objective differs across baselines; full details in Appendix~\ref{app:cluster-pipeline}.
\emph{Heuristic groupings.} We apply the published WebOrganizer Topic and Format taxonomies \citep{weborganizer} to our corpus at their native granularities. For every grouping, sampler axes are varied as in Section~\ref{sec:sampling}. Sample-wise selectors are out of scope for this baseline set; see Appendix~\ref{app:cluster-pipeline} for the scope statement.

\section{Results}
\label{sec:results}

We use the Results section to rule out two simpler explanations for HERMES's gain. The gain is not because HERMES is a better L1 grouping method: five 256-way methods sit at a plateau on standard compactness/mass-balance metrics and the two we have paired downstream are numerically tied (Section~\ref{sec:plateau}). It is not because finer granularity is universally better: at $L_{123}$, the within-bucket quality top-30\% rule's L12 advantage over quota-preserving coverage shrinks to a gap too small to interpret under single-seed evaluation, consistent with candidate competition (Section~\ref{sec:mechanism}). The reliable positive result is conditional: within a fixed DoReMi outer-weight family at granularity $L_{12}$, switching the Stage-2 sampler from max-entropy coverage to (corrected-reader) quality top-30\% improves Avg.\ by $+0.0253$ (z $+2.09$; Section~\ref{sec:findingA}). The broader outer-weight choice at this rule remains open: we report the highest-scoring configuration in our sweep in Section~\ref{sec:topline}, but we have not measured the Uniform + $L_{12}$ quality top-30\% counterpart and therefore cannot claim DoReMi is necessary for this gain. A paired Uniform vs.\ DoReMi $L_1$ characterisation is reported as an audit in Appendix~\ref{app:outerweight}.

\subsection{At \texorpdfstring{$L_1{=}256$}{L1=256}, grouping choice is not the source of gains}
\label{sec:plateau}

Before attributing any gain to HERMES, we first ask whether the gains are simply due to a better 256-way clustering. The answer is no: at $L_1{=}256$ the grouping axis sits on a plateau on selected compactness/mass-balance metrics (Avg cosine, $N_\text{eff}$, entropy), and the two grouping families we have paired downstream are numerically tied.

\paragraph{Intrinsic clustering metrics.} Five 256-bucket grouping methods evaluated on the same 1.16M-row fit set and 221k-row validation shard (Appendix~\ref{app:cluster-pipeline}); the baselines use the raw 1024-dim embeddings while HERMES applies its Learned Semantic Transform first. They sit tightly together (Table~\ref{tab:plateau}): mean cosine to centroid varies by less than $0.003$, effective cluster counts span 236--247, and entropies fall in $[5.46,5.51]$. Standard clustering criteria do not distinguish these methods at $L_1{=}256$.

\begin{table*}[t]
\centering
\caption{Intrinsic plateau: five learned grouping methods at $L_1{=}256$ are mutually indistinguishable on standard clustering metrics. Effective cluster count is the perplexity of the bucket-mass distribution, $N_\text{eff} = \exp(H)$ with $H = -\sum_k p_k \ln p_k$ and $p_k$ the fraction of documents in bucket $k$.}
\label{tab:plateau}
\small
\begin{tabularx}{\textwidth}{Xrrrr}
\toprule
Method & Eff.\ clusters & Entropy & Avg cos$\to$centroid & Recall@10 \\
\midrule
KMeans (sklearn) & 240.98 & 5.485 & 0.8740 & 0.356 \\
MiniBatchKMeans & 238.55 & 5.475 & 0.8731 & 0.345 \\
BisectingKMeans & 247.13 & 5.510 & 0.8709 & 0.285 \\
Plain RVQ ($c_1$) & 235.99 & 5.464 & 0.8719 & 0.348 \\
HERMES ($c_1$, ours) & 240.46 & 5.483 & 0.8739 & 0.355 \\
\bottomrule
\end{tabularx}
\end{table*}

\paragraph{Paired downstream training.} Two of the five have been paired against identical Uniform + L1 max-entropy: Uniform $\cdot$ HERMES-$256^3$ (Avg.\ $0.4155$, z $+1.048$) and Uniform $\cdot$ KMeans-256 (Avg.\ $0.4153$, z $+1.059$). They differ by $0.0002$ on Avg., numerically tied; the other three are an empirical question. With the L1 grouping axis ruled out, the remaining design axes are where HERMES's contribution must come from.

\paragraph{Auxiliary label-system diagnostics.} HERMES labels are moderately but not redundantly related to the WebOrganizer Topic, WebOrganizer Format, and Topic-over-Source taxonomies on a $259{,}255$-document intersection; finer HERMES granularity produces buckets that align more tightly with each reference family while remaining a distinct label basis (Appendix~\ref{app:label-relationship}). Qualitative inspection on a $4.8$M-document sample confirms this: all $256$ $L_1$ buckets are populated, and distinctive n-grams recover coherent topical or stylistic regions (e.g., books and publishing, music releases, recipes, video games, biomedicine, ophthalmology, astronomy), with $L_{12}$ children refining each $L_1$ region into narrower sub-regions (e.g., a Music parent splits into classic rock, music videos, MP3 pages, hip-hop, and gospel; Appendix~\ref{app:label-inspection}). A sparsity audit on the full $227$M-document annotated corpus separately rules out third-stage codebook collapse, verifies that every $L_1$ cell refines into a multi-child $L_{12}$ subtree, and shows that $L_{12}\!\to\!L_{123}$ refinement is concentrated in the higher-mass tail of $L_{12}$ cells (Appendix~\ref{app:sampling}, extended discussion).

\subsection{Finding A: Stage-2 rule choice matters at \texorpdfstring{$L_{12}$}{L12}}
\label{sec:findingA}

\begin{figure}[t]
\centering
\includegraphics[width=\columnwidth]{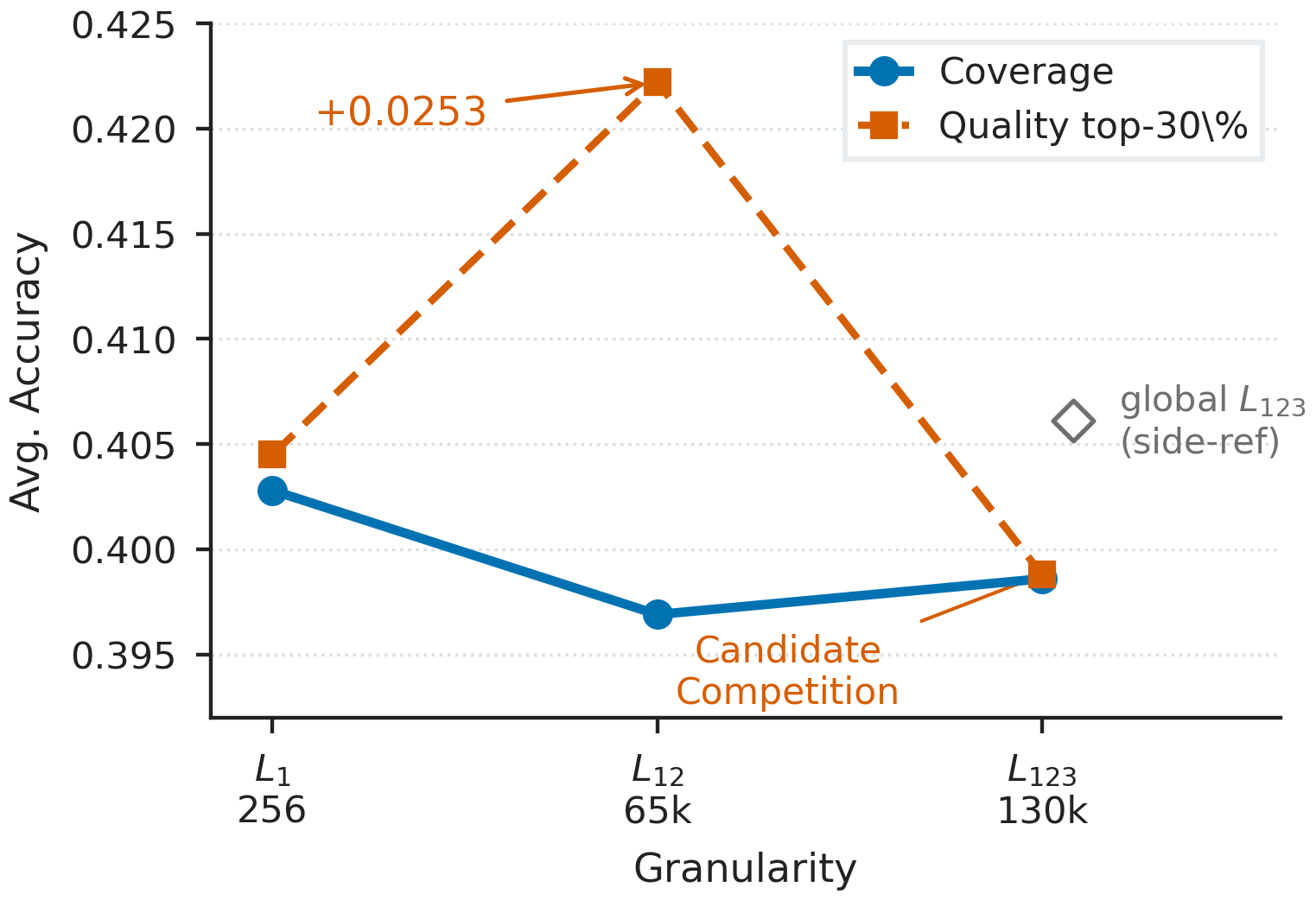}
\caption{Granularity arc under fixed HERMES codebook and DoReMi-L1 outer weights. Solid: quota-preserving coverage. Dashed: corrected-reader quality top-30\%. Open diamond: global $L_{123}$ max-entropy side-reference (quota-flattening). The $L_{12}$ quality gain collapses at $L_{123}$, consistent with candidate competition (Section~\ref{sec:mechanism}).}
\label{fig:granularity-arc}
\end{figure}

With the L1 clustering explanation ruled out (Section~\ref{sec:plateau}), we next isolate the Stage-2 sampler axis at a fixed granularity. Holding the L1 outer at DoReMi and the granularity at $L_{12}$, switching the Stage-2 sampler from max-entropy coverage to quality top-30\% (under the corrected FineWeb-Edu quality reader; Appendix~\ref{app:quality}) raises Avg.\ by $+0.0253$ (z $+2.09$). Only the Stage-2 rule changes between rows; the L1 weights, the granularity, and the HERMES-$256^3$ codebook are held fixed. The $L_{12}$ point is the peak of the granularity arc in Figure~\ref{fig:granularity-arc} (solid = quota-preserving coverage; dashed = quality top-30\%; both evaluated at DoReMi-L1 outer weights). The open diamond at $L_{123}\!=\!0.4061$ is the global $L_{123}$ max-entropy side-reference, which is a quota-flattening variant and not a quota-preserving Stage-2 contrast.

\begin{table}[t]
\centering
\caption{At fixed $L_{12}$ granularity under DoReMi L1 outer weights, switching the Stage-2 sampler from max-entropy coverage to quality top-30\% (corrected FineWeb-Edu reader) raises Avg.\ by $+0.0253$ (z $+2.09$). Each row is a single 1B/25B training seed; we discuss the single-seed limitation in Section~\ref{sec:limitations}.}
\label{tab:findingA}
\small
\begin{tabularx}{\columnwidth}{Xrr}
\toprule
Stage-2 sampler (DoReMi; granularity $L_{12}$) & Avg. & z \\
\midrule
$L_{12}$ max-entropy coverage & 0.3969 & $-0.458$ \\
$L_{12}$ quality top-30\% (corrected reader) & \textbf{0.4222} & $\mathbf{+1.628}$ \\
\bottomrule
\end{tabularx}
\end{table}

Quality top-30\% is a \emph{within-sub-bucket selection} rule: inside each $L_{12}$ ancestor, the sampler concentrates draws on the top-30\% of documents ranked by the FineWeb-Edu quality score, on size-proportional sub-bucket mass. Max-entropy is a \emph{coverage} rule: each sub-bucket is equalised regardless of natural size, with all documents inside eligible. The two rules therefore differ on two axes simultaneously, sub-bucket mass distribution (equal vs.\ size-proportional) and per-document eligibility (all docs vs.\ within-sub-bucket top-30\%), so $+0.0253$ is a combined Stage-2 rule contrast rather than a pure quality-ranking ablation; we do not attempt to isolate the two axes here. At $L_{12}$, sub-buckets are large enough (median pool 2{,}271 documents; Figure~\ref{fig:bucket-size}) that the within-bucket quality ranking is comparatively stable, so the quality top-30\% configuration extracts a meaningful signal that the coverage configuration misses. Whether the same Stage-2 advantage survives at finer granularities is the question Section~\ref{sec:mechanism} addresses.

\subsection{Finding B: the \texorpdfstring{$L_{12}$}{L12} advantage collapses at \texorpdfstring{$L_{123}$}{L123}}
\label{sec:mechanism}
\label{sec:findingB}

HERMES exposes a boundary on the granularity axis we call \emph{candidate competition}: as sub-buckets shrink, the within-bucket pool contracts, and per-bucket top-$k$ is no longer a stable proxy for the global top-$k$ it is meant to approximate. We measure this contraction directly on the sampler's realised selection set.

\begin{figure}[t]
\centering
\includegraphics[width=\columnwidth]{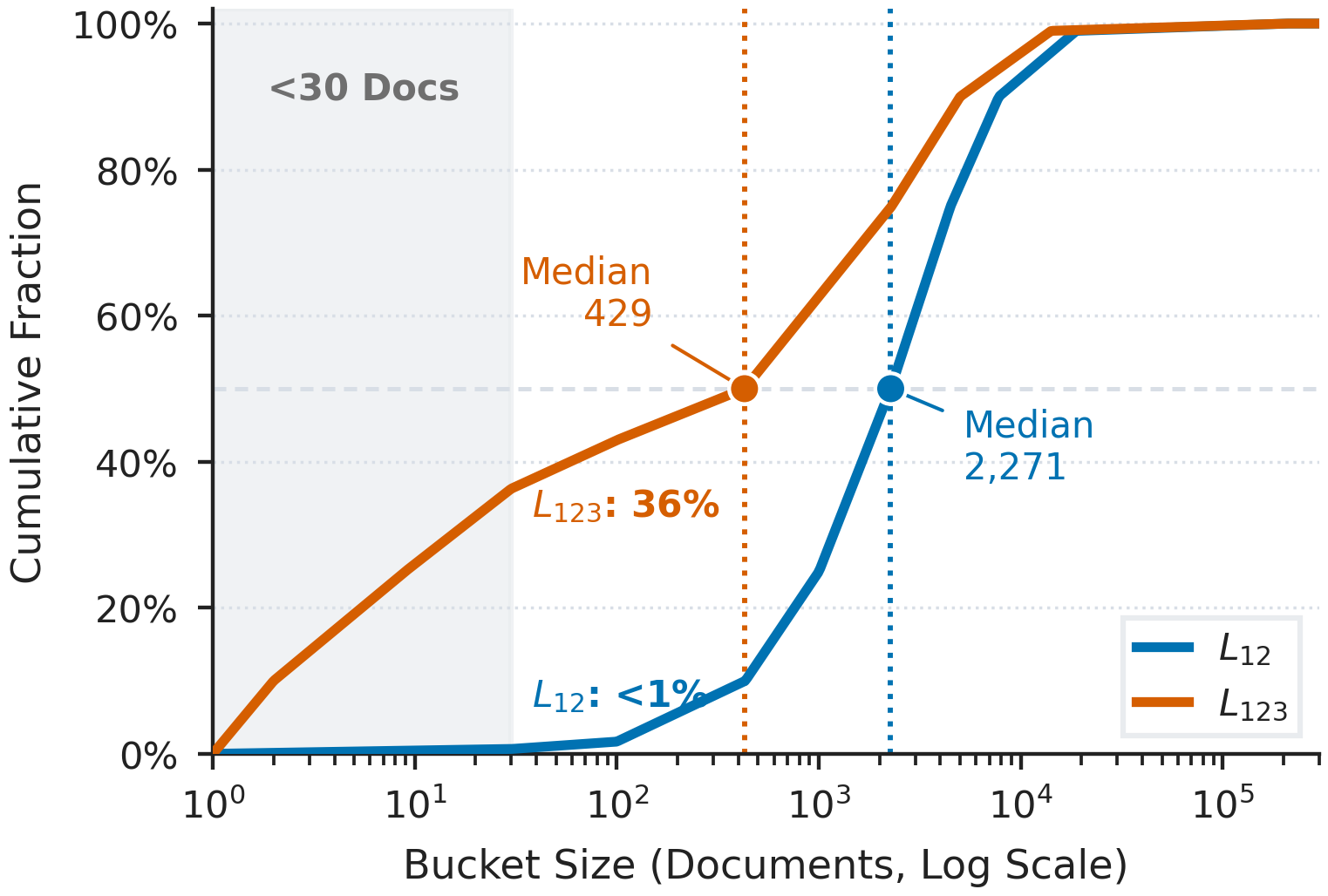}
\caption{Cumulative distribution of bucket sizes (log-scale x-axis) on the realised $L_{12}$ and $L_{123}$ selection sets. Refining to $L_{123}$ shifts the CDF leftward: the median pool falls from 2{,}271 to 429 documents, and 36\% of $L_{123}$ buckets contain fewer than 30 documents.}
\label{fig:bucket-size}
\end{figure}

The median candidate pool contracts $5.3\times$ from $L_{12}$ (2{,}271) to $L_{123}$ (429), and the contraction is robust across percentiles (Figure~\ref{fig:bucket-size}; full statistics in Appendix~\ref{app:sampling}). At this scale a per-sub-bucket top-30\% rule reduces to ranking inside pools of ${\sim}429$ documents, a regime in which the within-sub-bucket score is a less reliable proxy for the global ranking. Max-entropy at $L_{123}$ has no such dependency. The mechanism is therefore consistent with: at fine granularity, the L12 Stage-2 advantage of quality top-30\% over quota-preserving coverage should diminish.

The leaderboard supplies a direct test. Holding DoReMi-L1 outer weights fixed and pairing both Stage-2 rules in their \emph{quota-preserving} forms (Stage-1 L1 weights preserved exactly), the $+0.0253$ L12 advantage of quality top-30\% over coverage collapses at $L_{123}$: quality top-30\% reaches $0.3988$ and per-L1 $L_{123}$ local random coverage reaches $0.3986$, a $\Delta=+0.0002$ that is too small to interpret under single-seed evaluation. The L12 Stage-2 advantage of quality top-30\% over coverage thus loses its measurable edge at $L_{123}$, consistent with the candidate-competition account in which within-bucket quality ranking is no longer measurable over coverage once median pools contract to ${\sim}429$ documents. A separate quota-flattening variant (global $L_{123}$ max-entropy, $0.4061$) is a side-reference (Appendix~\ref{app:leaderboard}) that differs from $L_{123}$ quality top-30\% on both the Stage-2 rule \emph{and} quota preservation, so we do not read it as a fair Stage-2 contrast. We frame this as evidence consistent with candidate competition rather than causal isolation; the full Stage-2 rule $\times$ granularity table is in Appendix~\ref{app:findingB-table}, and ``only-large-sub-buckets''/``synthetic-shrinkage'' controls are noted in Section~\ref{sec:limitations}.

\subsection{Synthesis: the rank-1 configuration is L12 quality top-30\% under DoReMi-L1}
\label{sec:topline}

We report the highest-scoring configuration in our sweep alongside its closest neighbours (full leaderboard in Table~\ref{tab:leaderboard}, Appendix~\ref{app:leaderboard}); we do not claim to have identified the optimal configuration across the full design space.

Rank 1 (HERMES + DoReMi-L1 + $L_{12}$ quality top-30\% under the corrected reader, Avg.\ $0.4222$) applies the DoReMi L1 weights and then, within each $L_{12}$ sub-bucket, concentrates draws on the top-30\% of documents ranked by the FineWeb-Edu quality score: the DoReMi L1 quota is preserved, and within-bucket selection extracts a meaningful quality signal at a granularity where pools are still large enough for the ranking to be stable (Section~\ref{sec:mechanism}). Ranks 2--4 sit at the $L_1{=}256$ plateau: HERMES, KMeans, and WebOrganizer-Format under Uniform + L1 max-entropy land within ${\sim}0.002$ Avg.\ of one another, consistent with Section~\ref{sec:plateau}. \textbf{What we do not claim.} We have not run a Uniform + $L_{12}$ quality top-30\% configuration under the corrected reader, so the outer-weight choice at the rank-1 Stage-2 rule cannot be adjudicated from the leaderboard alone; in particular, we cannot conclude that DoReMi is necessary for the rank-1 Avg. The paired Uniform vs.\ DoReMi $L_1$ characterisation (Appendix~\ref{app:outerweight}) shows that the choice between our two $O(1)$-proxy outer weights is not free at $L_1$, but a fair comparison at $L_{12}$ quality top-30\% would require an additional 1B/25B run.

\section{Discussion}
\label{sec:discussion}

The contribution is the substrate, not the clustering algorithm: HERMES sits at a plateau with four 256-bucket clusterers on standard compactness/mass-balance metrics and is downstream-matched to KMeans within $0.0002$ Avg.\ (Section~\ref{sec:plateau}). What HERMES adds is a single offline annotation in which granularity becomes a free variable, exposed as a prefix length. In a flat KMeans pipeline a granularity sweep is a sequence of independent clusterings; in HERMES the same hierarchy is read at different prefix lengths, which is what makes the granularity-by-sampler interaction (Section~\ref{sec:mechanism}) cleanly measurable. More structurally, HERMES turns the corpus-labeling axis (previously a binary choice between coarse fixed labels and per-document scoring) into a continuous granularity dial the practitioner sets by prefix length, paying per-document annotation cost once and trading resolution at sample time.

\section{Conclusion}
\label{sec:conclusion}

HERMES is a data-derived multi-granularity corpus-labeling substrate built from a Learned Semantic Transform and a 3-stage RVQ: one annotation, prefix-readable granularity from $L_1{=}256$ to $L_{123}{\sim}130$k. At $L_1{=}256$, HERMES sits at a plateau with the KMeans family on standard clustering metrics and is downstream-matched within $0.0002$ Avg., so the contribution is the substrate, not the clusterer. At $L_{12}$ under DoReMi-L1, switching the Stage-2 sampler from max-entropy coverage to corrected-reader quality top-30\% raises Avg.\ by $+0.0253$ (z $+2.09$; Finding A). The advantage collapses at $L_{123}$ (quality vs.\ per-L1 local random coverage: $0.3988$ vs.\ $0.3986$; Finding B), consistent with candidate competition: a $5.3\times$ median pool contraction (2{,}271 $\to$ 429) leaves the within-bucket ranking signal no longer measurable over coverage. Pre-training data design becomes a label-system question: granularity and the Stage-2 sampler are jointly determined axes on the substrate, not independent hyperparameters. Promising next directions are (a) deeper or wider RVQ hierarchies as corpus size grows, with the useful depth bounded by candidate competition; (b) jointly learned outer weights at multiple prefix lengths; and (c) soft within-bucket coverage signals, including intrinsic per-bucket priors such as the RVQ reconstruction error already produced by the substrate, as an alternative to hard top-$k$ selection on external scorers.

\section*{Limitations}
\label{sec:limitations}

\textbf{Scale.} All conclusions are at 1B parameters and 25B tokens on one internal corpus; whether the interaction persists at 7B/300B is not validated here. Sub-bucket-size shrinkage is a structural property of residual quantization rather than a small-scale artefact, so we expect the qualitative direction to survive, but the granularity at which the candidate pool collapses may shift with scale \citep{scaling-laws}.

\textbf{Encoder.} A single frozen encoder produces the input geometry; different encoders may shift the realised $L_{123}$ cardinality and the threshold at which candidate competition becomes dominant.

\textbf{Outer-weight axis is under-sampled.} Appendix~\ref{app:outerweight} pairs Uniform vs DoReMi at L1 + L1 max-entropy only; our DoReMi weights are learned once at L1 and reused at $L_{12}$ and $L_{123}$ without re-learning (the proxy budget makes natively training DoReMi proxies at finer granularity impractical, and regression-style alternatives are infeasible at $K=256$--$130$k). Whether learned outer weights at the granularity actually used by Stage-2 close or invert the gap is open.

\textbf{Outer weight at rank-1 Stage-2 rule is under-tested.} Our highest-scoring configuration (Section~\ref{sec:topline}) uses DoReMi outer weights with $L_{12}$ quality top-30\% under the corrected reader, but we did not run the parallel Uniform + $L_{12}$ quality top-30\%. Without this counterfactual, we cannot claim DoReMi is necessary for the rank-1 Avg., nor that DoReMi is the optimal outer-weight family for this Stage-2 rule. The Stage-2 contrast we report (Section~\ref{sec:findingA}, within DoReMi) is a controlled within-slice comparison and is not affected by this gap.

\textbf{Within-sub-bucket scoring function.} The quality samplers use the FineWeb-Edu quality score field released with the FineWeb corpus \citep{fineweb-edu}. Whether the sign of Finding A and the advantage collapse at $L_{123}$ in Finding B hold under perplexity-based or other classifier-based scoring families is a direct follow-up; scores whose distribution differs sharply from FineWeb-Edu's right tail may interact with the small-pool regime differently.

\textbf{Mechanism evidence strength.} The $5.3\times$ median sub-bucket shrinkage is a direct measurement, and the L12 advantage collapse at $L_{123}$ under the quota-preserving pair (Section~\ref{sec:findingB}) is a paired training contrast in the direction consistent with the candidate-competition account. We treat this as evidence consistent with the mechanism rather than causal isolation. Two controls that would tighten the claim, the ``only-large-sub-buckets'' and ``synthetic-shrinkage'' variants, are future work.

\textbf{Hierarchy depth and codebook size.} We sweep granularity by prefix length on a single $L=3$, $K=256$ codebook chosen for the ${\sim}50$M-document corpus. Deeper hierarchies (larger $L$) or wider codebooks (larger $K$, possibly non-uniform across stages) are natural extensions. The candidate-competition account suggests diminishing returns once deeper prefixes push the median sub-bucket below the regime where within-bucket ranking is stable (${\sim}429$ documents at $L_{123}$ here); larger corpora should support deeper hierarchies before this floor is hit. The framework's extensibility via RVQ-style codebook design, including deeper hierarchies (larger $L$) and non-uniform per-stage $K$, is, we argue, a positive selling point of the substrate rather than a caveat. Systematic depth-by-corpus-size sweeps are left to future work.

\section*{Artifact Release}
We release the source code implementing the HERMES annotation pipeline (LST training, RVQ training, and inference). An anonymized artifact submission accompanies this paper.

\section*{Ethical Considerations}
The pre-training corpus is an internal ${\sim}50$M-document collection that has been pre-filtered with public quality classifiers; we treat it as a fixed source distribution and study \emph{sampling}, not curation. No additional PII filtering is performed by our work: we inherit the upstream classifier's coverage. The 20 ranked 1B/25B runs reported here total approximately 100 GPU-days on 8 GPUs at FP32; we did not run extra ablations beyond what is reported. The corpus and encoder are internal, which limits direct reproduction; the source code release described above allows others to apply the pipeline to their own corpus and encoder.

\bibliography{custom}

\clearpage
\appendix

\FloatBarrier
\section{Capacity Ablation: L1 Codebook Size}
\label{app:capacity}

To isolate the role of the coarsest stage, we hold stages 2 and 3 at $K=256$ and vary the L1 codebook size $K\in\{32,64,128,256\}$. All four runs share the same encoder, training recipe, and Uniform + L1 max-entropy sampler.

\begin{figure}[t]
\centering
\includegraphics[width=\columnwidth]{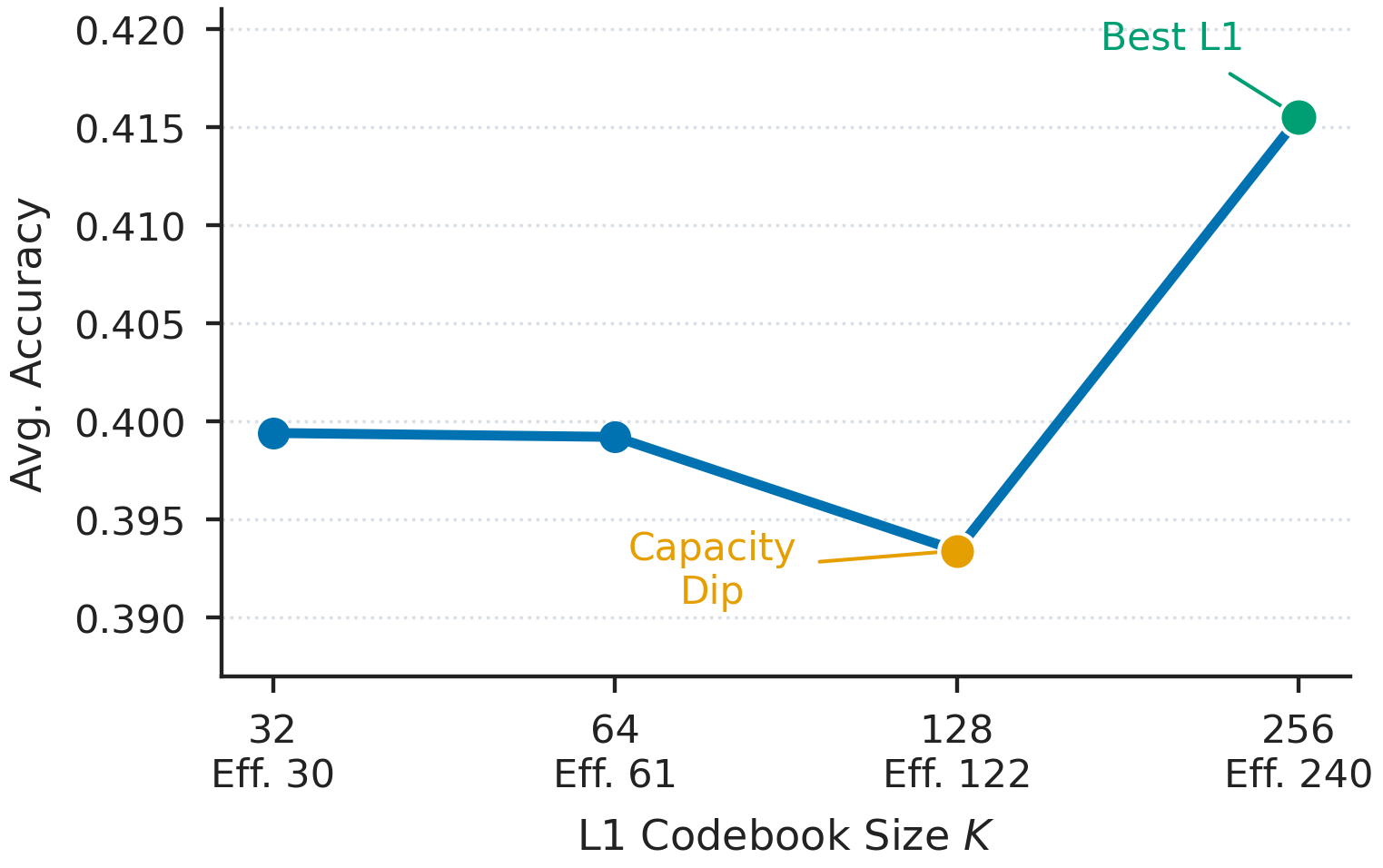}
\caption{HERMES L1 capacity ablation. Holding stages 2--3 fixed, the L1 capacity sweep of 16-task Avg.\ accuracy is non-monotone: $K=128$ is the trough, while $K=256$ is the best L1 setting in this study. X-axis tick labels show both the nominal codebook size $K$ and the realised effective cluster count $N_\text{eff}$ at that $K$ (saturating near $240$ at $K=256$; see Table~\ref{tab:plateau} for the $N_\text{eff}$ definition).}
\label{fig:capacity-ablation}
\end{figure}

We read this through L1's role as a capacity gate. \emph{Below the gate} ($K=32,64$), L1 under-resolves the corpus; stages 2--3 absorb dimensions of variation L1 should have captured, so each L1 bucket is internally heterogeneous and the sampler falls back to a near-uniform draw. \emph{At the trough} ($K=128$), L1 acts non-trivially but is mis-aligned with corpus structure; the sampler can no longer fall back to a uniform draw, but its reweighting points in the wrong direction. \emph{At $K=256$}, the effective cluster count saturates near nominal (240/256), L1 alone carries the partition, and Avg.\ jumps. The shape is not ``more buckets is always better''; it is ``L1 must be large enough that bucket-level reweighting and corpus structure are aligned.'' This ablation is a capacity dimension within a single grouping family and does not contradict the $L_1{=}256$ plateau across grouping methods (Section~\ref{sec:plateau}).

\section{Operational Definitions of Granularity}
\label{app:granularities}

We use three granularity levels, each defined by a prefix of the HERMES code $(c_1,c_2,c_3)$. L1 saturates its nominal capacity. L12 is essentially fully populated (99.8\% of nominal $256^2$ joint codes are observed). L123 deviates substantially: only 129{,}955 of $256^3\approx 16.8\text{M}$ nominal joint codes are realized, a roughly two-orders-of-magnitude gap. We attribute this to natural sparsity of stage-3 RVQ residuals on this corpus: once the first two stages have absorbed the dominant directions of variation, the third-stage residuals concentrate on a thin manifold of the joint code space. L123 is operationally defined as ``the prefix length that realizes approximately 130k buckets,'' and it is in this regime that candidate competition takes hold.

\section{Full Leaderboard}
\label{app:leaderboard}

\begin{table*}[t]
\centering
\caption{Full leaderboard sorted by Avg.\ (descending), with z standardised across all 20 ranked 1B/25B checkpoints (14 main rows + 6 HERMES capacity-ablation rows). All rows use a single training seed. ``Corrected'' marks runs that use the corrected FineWeb-Edu quality reader (Appendix~\ref{app:quality}).}
\label{tab:leaderboard}
\small
\begin{tabularx}{\textwidth}{rXllrr}
\toprule
Rank & Configuration & L1 outer & Stage-2 sampler & Avg. & z \\
\midrule
1 & HERMES + $L_{12}$ quality top-30\% (corrected) & DoReMi & $L_{12}$ quality top-30\% & 0.4222 & $+1.628$ \\
2 & HERMES-$256^3$ + L1 max-entropy (ours) & Uniform & L1 max-entropy & 0.4155 & $+1.048$ \\
3 & KMeans-256 + L1 max-entropy & Uniform & L1 max-entropy & 0.4153 & $+1.059$ \\
4 & WebOrganizer-Format & Uniform on Format & random within Format & 0.4134 & $+0.949$ \\
5 & WebOrganizer-Topic + Topic DRO & DoReMi on Topic & random within Topic & 0.4082 & $+0.508$ \\
6 & WebOrganizer-Topic + Topic max-entropy & Uniform on Topic & random within Topic & 0.4079 & $+0.469$ \\
7 & HERMES + global $L_{123}$ max-entropy coverage & DoReMi & global $L_{123}$ max-entropy & 0.4061 & $+0.317$ \\
8 & HERMES + L1-local quality top-30\% (corrected) & DoReMi & L1-local quality top-30\% & 0.4045 & $+0.114$ \\
9 & WebOrganizer-Format + Format DRO & DoReMi on Format & random within Format & 0.4029 & $+0.128$ \\
10 & HERMES-$256^3$ + L1 max-entropy (ours) & DoReMi & L1 max-entropy & 0.4028 & $+0.048$ \\
11 & HERMES cap-32 + L1 max-entropy & Uniform & L1 max-entropy & 0.3994 & $-0.312$ \\
12 & HERMES cap-64 + L1 max-entropy & Uniform & L1 max-entropy & 0.3992 & $-0.362$ \\
13 & HERMES + $L_{123}$ quality top-30\% (corrected) & DoReMi & $L_{123}$ quality top-30\% & 0.3988 & $-0.294$ \\
14 & HERMES + per-L1 $L_{123}$ local random coverage & DoReMi & per-L1 $L_{123}$ local random & 0.3986 & $-0.302$ \\
15 & HERMES + $L_{12}$ max-entropy coverage & DoReMi & $L_{12}$ max-entropy & 0.3969 & $-0.458$ \\
16 & HERMES cap-128 + L1 max-entropy & Uniform & L1 max-entropy & 0.3934 & $-0.747$ \\
17 & KMeans-256 + L1 max-entropy & DoReMi & L1 max-entropy & 0.3889 & $-1.147$ \\
18 & HERMES cap-64 + L1 max-entropy & DoReMi & L1 max-entropy & 0.3887 & $-1.163$ \\
19 & HERMES cap-128 + L1 max-entropy & DoReMi & L1 max-entropy & 0.3886 & $-1.181$ \\
20 & HERMES cap-32 + L1 max-entropy & DoReMi & L1 max-entropy & 0.3798 & $-1.885$ \\
\bottomrule
\end{tabularx}
\end{table*}

The top of the table is occupied by mechanism-aligned configurations: at fixed DoReMi L1 outer weights, $L_{12}$ quality top-30\% under the corrected reader (rank 1), and the Uniform + L1 max-entropy family (ranks 2--4). WO-Topic under both outer-weight families (ranks 5, 6) and global $L_{123}$ max-entropy coverage (rank 7) sit just below. The middle tier corresponds jointly to the $L_1{=}256$ grouping plateau (Section~\ref{sec:plateau}) and the outer-weight paired plateau (Appendix~\ref{app:outerweight}): HERMES, KMeans, and WebOrganizer-Format under Uniform + L1 max-entropy are within roughly $0.002$ Avg.\ of one another (ranks 2--4); the same grouping families under DoReMi land $0.9$--$2.1$ z lower (ranks 9, 10, 17), with the WO-Topic pair (ranks 5, 6) being the one too small to interpret under single-seed evaluation. The bottom tier is populated by $L_{123}$-granularity rows under DoReMi (ranks 13, 14, 15, which sit within $0.002$ of one another and below the $L_1$ plateau) and capacity-starved HERMES variants under both Uniform (ranks 11, 12, 16) and DoReMi (ranks 18, 19, 20). The L12 quality top-30\% advantage at rank 1 collapses to noise at $L_{123}$ (rank 13 quality top-30\% Avg.\ 0.3988 vs.\ rank 14 per-L1 $L_{123}$ local random coverage Avg.\ 0.3986), consistent with the candidate-competition account in Section~\ref{sec:mechanism}.

\section{HERMES Training Details}
\label{app:hermes-train}

\paragraph{Encoder.} An internally trained 1024-dim sentence-level encoder, kept frozen throughout HERMES training. Embeddings are read once per document; no fine-tuning at this stage.

\paragraph{Optimizer and schedule.} AdamW, learning rate $3\times 10^{-4}$, weight decay $1\times 10^{-4}$, 10 epochs, 8-GPU distributed data-parallel (DDP), batch size 1024, single-precision (FP32) throughout.

\paragraph{Codebook initialization and EMA.} EMA codebook updates with dead-code threshold 2 and $k$-means initialization (10 iterations); stage $k$'s codebook is learned in the residual space left by stages $1,\ldots,k-1$, so prefix codes form a strict hierarchy.

\paragraph{LST loss equations.} LST is trained jointly with the RVQ codebooks under three named losses, following the rotate-then-quantize lineage of \citet{opq}. \emph{Pairwise structure preservation} keeps cosine geometry between raw and transformed embeddings; over $M=2048$ random pairs $(i,j)$ per minibatch,
\[
\mathcal{L}_\text{struct} = \tfrac{1}{M}\sum_{(i,j)}\bigl(\cos(e_i,e_j) - \cos(h_i,h_j)\bigr)^2.
\]
\emph{Quantization-aware reconstruction} couples LST to the codebooks. Letting $\hat h_i = \sum_{k=1}^{3} q_k(h_i)$ denote the RVQ reconstruction (projected to the unit sphere), over a minibatch of size $B$,
\[
\mathcal{L}_\text{quant} = \tfrac{1}{B}\sum_{i=1}^{B}\|\hat h_i - h_i\|_2^2.
\]
\emph{Orthogonality} prevents representation collapse: $\mathcal{L}_\text{ortho} = \|W^\top W - I\|_F^2$, combined with a per-step SVD projection onto the orthogonal manifold (writing $W=U\Sigma V^\top$, set $W\leftarrow UV^\top$). The total LST objective is
\begin{align*}
\mathcal{L} = {} & \lambda_\text{struct}\mathcal{L}_\text{struct} + \lambda_\text{quant}\mathcal{L}_\text{quant} \\
& + \lambda_\text{ortho}\mathcal{L}_\text{ortho} + \mathcal{L}_\text{commit},
\end{align*}
with $\lambda_\text{struct}=\lambda_\text{quant}=1.0$ and $\lambda_\text{ortho}=0.1$. The per-stage commitment term is
\[
\mathcal{L}_\text{commit} = \beta\,\sum_{k=1}^{L} \bigl\|\,\mathrm{sg}(q_k) - r_k\,\bigr\|_2^2,
\]
with $\beta = 0.25$ and $\mathrm{sg}(\cdot)$ the stop-gradient operator.

\paragraph{Inference.} Implementation builds on the open-source \emph{vector-quantize-pytorch} library \citep{vqpytorch}; we annotate the entire ${\sim}50$M-document corpus once and reuse the same annotation across all sampler experiments.

\section{Annotation Pipeline for Unsupervised Clustering Baselines}
\label{app:cluster-pipeline}

All unsupervised clustering baselines reported in Section~\ref{sec:plateau} and Table~\ref{tab:plateau} (KMeans, MiniBatchKMeans, BisectingKMeans, Plain RVQ) share an identical two-stage annotation procedure that differs from HERMES only in the clustering objective; HERMES's own training is described in Appendix~\ref{app:hermes-train}.

\paragraph{Inputs.} The same ${\sim}50$M-document corpus as HERMES, with the same precomputed 1024-dim sentence-level embeddings $e_i$. No embedding is recomputed.

\paragraph{Stage 1: Centroid fitting on a held-out subsample.} All baselines fit on a fixed subsample of 1{,}158{,}563 embeddings (the same set across methods); a separate 221{,}476-embedding validation shard reports intrinsic metrics for Table~\ref{tab:plateau}. Shared settings: $K=256$ centroids, embeddings un-normalised at fit time and L2-normalised at assignment, a fixed random seed, and up to $100$ update iterations. KMeans, MiniBatchKMeans, and BisectingKMeans all use scikit-learn's default initialisation and restart policies for each algorithm; the only non-default we set is MiniBatchKMeans's minibatch size at $8192$. Plain RVQ uses EMA codebook updates on the same 1.16M-row sample, \emph{without} the Learned Semantic Transform; this is what ``plain'' means in our naming.

\paragraph{Stage 2: Full-corpus annotation.} Saved centroids are L2-normalised once and applied to every document in the full corpus by cosine assignment, $\arg\max_k\,\tilde e_i^\top \tilde c_k$ (argmin Euclidean is supported but not used). Annotation runs once per method and is reused across every sampler ablation in this paper, mirroring HERMES's reuse (Appendix~\ref{app:hermes-train}).

\paragraph{Why this licenses the plateau claim.} The 1.16M-row fit set, 1024-dim raw embedding, $K=256$, normalisation convention, iteration budget, and random seed are identical across KMeans, MiniBatchKMeans, BisectingKMeans, and Plain RVQ; the only thing that differs is the clustering objective. This is what licenses the Section~\ref{sec:plateau} claim that intrinsic Avg.\ cosine spreads by less than $0.003$ across all five 256-bucket methods (KMeans $0.8740$, MiniBatchKMeans $0.8731$, BisectingKMeans $0.8709$, Plain RVQ-$c_1$ $0.8719$, HERMES-$c_1$ $0.8739$): they are not measured on different corpora or under different normalisation conventions.

\paragraph{Sample-wise selectors are out of scope for this baseline set.} DSIR \citep{dsir}, QuRating \citep{qurating}, MATES \citep{mates}, LESS \citep{less}, DataInf \citep{datainf}, SampleMix \citep{samplemix}, and QuaDMix \citep{quadmix} score documents individually rather than reweighting groups. At the $50$M-document $\times$ 1B-parameter $\times$ 25B-token scale we target they are costly to score, tune, and validate, and most do not expose an explicit mixture-control interface that a granularity-aware sampler could plug into. A head-to-head against per-sample selectors at smaller regimes is a natural follow-up.

\section{Sampling Diagnostics}
\label{app:sampling}

We summarise the bucket-size statistics that underpin the candidate-competition diagnostic. Per-granularity statistics are computed on the actual sampled set used by each sampler (${\sim}68$M selected documents under DoReMi outer weights for the L12 and L123 samplers; sampling is with replacement, so the sampled set exceeds the ${\sim}50$M source corpus), not on the raw 50M-document corpus, so they reflect the candidate pool that within-sub-bucket selection sees.

\begin{table}[t]
\centering
\caption{Per-granularity bucket statistics on realised selection sets. The $L_{12}\to L_{123}$ median shrinks $5.3\times$.}
\label{tab:sampling}
\small
\begin{tabularx}{\columnwidth}{Xrrr}
\toprule
Granularity & Active buckets & Median & Documents \\
\midrule
L1 & 256 & n/a & ${\sim}50$M \\
L12 & 65{,}408 & 2{,}271 & ${\sim}68$M \\
L123 & 129{,}955 & 429 & ${\sim}68$M \\
\bottomrule
\end{tabularx}
\end{table}

Full bucket-size distributions on log scale show the $L_{12}\to L_{123}$ bulk shifting roughly an order of magnitude leftward, with the $L_{123}$ lower tail developing a heavy concentration of sub-buckets containing only a handful of documents: the regime in which any within-sub-bucket ranking criterion reduces to a near-uniform draw. ``Active buckets'' counts sub-buckets receiving at least one document; for $L_{123}$ this is 129{,}955 out of nominal $256^3$.

\paragraph{Further sparsity statistics on the full annotated corpus.} On the full $227{,}447{,}180$-document annotated set, we additionally compute the effective cell count $N_\text{eff}=\exp(H)$, the Gini coefficient over per-cell document counts, the top-$k$ for $50\%$ cumulative mass, and per-cell P10/P90 (Table~\ref{tab:sparsity-extended}). We also compute, for every parent at $L_1$ and $L_{12}$, whether a single child carries more than $90\%$ of the parent's mass. Per-cell medians are reported in Table~\ref{tab:sampling} above (and Figure~\ref{fig:bucket-size}); we do not duplicate them here.

\begin{table}[t]
\centering
\caption{Per-prefix sparsity statistics on the full $227$M-document annotated corpus, complementing Table~\ref{tab:sampling}.}
\label{tab:sparsity-extended}
\footnotesize
\setlength{\tabcolsep}{3pt}
\begin{tabularx}{\columnwidth}{@{}>{\raggedright\arraybackslash}X r r r@{}}
\toprule
Statistic & $L_1$ & $L_{12}$ & $L_{123}$ \\
\midrule
Nominal cells              & 256   & 65{,}536           & $1.68{\times}10^{7}$ \\
$N_\text{eff}{=}\exp(H)$   & 240.5 & 40{,}693           & 46{,}007 \\
Gini over cell mass        & 0.20  & 0.52               & 0.73 \\
Top-$k$ for $50\%$ mass    & 95    & 11{,}215           & 12{,}268 \\
P10 per cell               & 534k  & 433                & 2 \\
P90 per cell               & 1.27M & 7{,}798            & 5{,}017 \\
${>}90\%$-dominant parents & 0/256 & 57{,}718/65{,}408  & --- \\
\bottomrule
\end{tabularx}
\end{table}

\paragraph{What this rules out.} Effective cell count rises monotonically ($240.5\to 40{,}693\to 46{,}007$): the third RVQ stage adds effective discrimination above $L_{12}$ even though only $0.77\%$ of nominal $L_{123}$ cells are populated. The Gini coefficient rises with granularity ($0.20\to 0.52\to 0.73$): mass becomes Zipfian as a fixed corpus is distributed over more cells, but no cell at any level carries more than $\sim$$1\%$ of the corpus.

\paragraph{Parent-child refinement.} Zero of the $256$ $L_1$ cells have a single $L_{12}$ child carrying more than $90\%$ of the parent's mass: every $L_1$ cell refines into a genuine multi-child subtree. At $L_{12}\!\to\!L_{123}$, $57{,}718$ of $65{,}408$ $L_{12}$ cells ($88\%$) have a dominant ($\!>$$90\%$) $L_{123}$ child. This is expected given that the median $L_{12}$ bucket carries only $2{,}271$ documents (Table~\ref{tab:sampling}): most $L_{12}$ cells do not have enough mass to populate multiple $L_{123}$ children with meaningful weight. The remaining ${\sim}7{,}700$ $L_{12}$ cells (the higher-mass tail) account for the bulk of $L_{123}$ refinement; this pattern supports the Section~\ref{sec:mechanism} regime, in which sub-buckets containing meaningful refinement coexist with low-mass tails.

\section{Diagnostic and Side-Reference Stage-2 Samplers}
\label{app:sampling-variants}

Beyond the four main Stage-2 samplers defined in Section~\ref{sec:sampling}, we use one diagnostic and two side-reference variants on the HERMES substrate. \emph{L1-local quality top-30\%} (diagnostic) selects, within each $L_1$ ancestor, the top-30\% by FineWeb-Edu quality on size-proportional $L_{12}$ mass; this row anchors the $L_1$ point of the granularity arc (Appendix~\ref{app:arc}). \emph{Per-L1 $L_{123}$ local random coverage} (side-reference, quota-preserving) draws $L_{123}$ sub-buckets uniformly within each $L_1$ ancestor, preserving the Stage-1 L1 quota exactly; this is the coverage row that pairs with $L_{123}$ quality top-30\% in Finding B (Section~\ref{sec:findingB}). \emph{Global $L_{123}$ max-entropy} (side-reference, quota-flattening) flattens all ${\sim}130$k $L_{123}$ sub-buckets corpus-wide; it inadvertently re-shapes the L1 mass towards L1 cells that split into many $L_{123}$ children, so it is not a quota-preserving Stage-2 contrast. All three variants appear in the leaderboard (Appendix~\ref{app:leaderboard}); only the per-L1 local random row is used as the coverage end of the quota-preserving Finding B pair.

\section{Quality Field Sanity}
\label{app:quality}

\textbf{(a) Field presence.} On a 2{,}000-row sanity scan of the corpus metadata, the nested path containing the FineWeb-Edu quality score was populated in 2000/2000 rows, while a top-level fallback field was populated in 0/2000. The corrected reader prefers the nested path and falls back to the top-level only for forward compatibility. All quality top-30\% rows discussed in Findings A and B (Sections~\ref{sec:findingA},~\ref{sec:findingB}) and listed in the leaderboard (Appendix~\ref{app:leaderboard}) use the corrected reader.

\textbf{(b) Threshold sanity number.} A dry run of the corrected reader over 556{,}976 quality values yields a global 70th-percentile $\tau_\text{global}=1.374$, statistically indistinguishable from the nested-only $p_{70}=1.380$ measured directly on the sanity scan. $\tau_\text{global}$ is a sanity number only; the within-sub-bucket quality samplers compute their thresholds \emph{per sub-bucket} via a bounded heap that keeps the local top-30\% within each sub-bucket, so the actual threshold $\tau_B$ varies with $B$ and is, on heavy-tailed sub-buckets, very different from $\tau_\text{global}$.

\section{Capability Sub-task List}
\label{app:olmes}

We evaluate on 16 capability sub-tasks grouped into four ability families. The six Basic Skills probes are taken from the OLMES capability suite \citep{olmes}; the remaining ten are standard public benchmarks.

\textbf{Basic Skills (6 tasks):} surface-level reasoning and string manipulation probes from OLMES \citep{olmes}: arithmetic, coding, common knowledge, logical reasoning, pattern recognition, and string operations.

\textbf{Science QA (5 tasks):} factual/scientific knowledge: ARC-Easy and ARC-Challenge \citep{arc}, SciQ \citep{sciq}, PIQA \citep{piqa}, and the DB-QA subset of LAB-Bench \citep{labbench}.

\textbf{Language Modeling (1 task):} sentence-level commonsense continuation: HellaSwag \citep{hellaswag}.

\textbf{Others (4 tasks):} open-domain and social knowledge: CommonsenseQA \citep{csqa}, Jeopardy (a quiz-show open-domain task), Natural Questions \citep{nq}, and Social IQA \citep{siqa}.

All sub-tasks are run in their standard pretrain configuration; per-task metric is accuracy. Avg.\ is the family-equal macro-mean defined in Section~\ref{sec:eval}.

\section{Granularity Arc Under Quality Top-30\%}
\label{app:arc}

The candidate-competition mechanism makes a directional suggestion along the granularity axis when Stage-2 is held at per-sub-bucket FineWeb-Edu quality top-30\% under DoReMi-L1 outer weights: the arc should peak at the intermediate granularity where buckets are large enough for the within-bucket quality ranking to be stable, with the $L_{12}\!\to\!L_{123}$ leg falling once the sub-bucket pool contracts. The three corrected-reader rows we have are consistent with this picture: the arc rises from $L_1$ to $L_{12}$, then falls back to ${\sim}0.399$ at $L_{123}$, where it converges with the quota-preserving $L_{123}$ per-L1 local random coverage row ($0.3986$; Appendix~\ref{app:leaderboard}). The candidate-competition account is therefore supported by the $L_{12}\!\to\!L_{123}$ drop in the quality top-30\% line, not by a quality-vs-coverage sign flip at $L_{123}$.

\begin{table}[t]
\centering
\caption{Granularity arc under DoReMi-L1 outer weights and within-sub-bucket quality top-30\% (corrected FineWeb-Edu reader). The arc peaks at $L_{12}$; the $L_{12}\!\to\!L_{123}$ drop is consistent with the candidate-competition account in Section~\ref{sec:mechanism}.}
\label{tab:arc}
\small
\begin{tabularx}{\columnwidth}{Xrr}
\toprule
Granularity (DoReMi; quality top-30\% Stage-2, corrected) & Avg. & z \\
\midrule
$L_1$ (L1-local quality top-30\%) & 0.4045 & $+0.114$ \\
$L_{12}$ ($\sim$65k buckets) & \textbf{0.4222} & $\mathbf{+1.628}$ \\
$L_{123}$ ($\sim$130k buckets) & 0.3988 & $-0.294$ \\
\bottomrule
\end{tabularx}
\end{table}

The peak at $L_{12}$ and the drop at $L_{123}$ are consistent with the candidate-competition account: at $L_{123}$, the median candidate pool is $429$ documents (Appendix~\ref{app:sampling}), below the regime where within-sub-bucket ranking is stable. A non-decreasing arc, or a coarse-end collapse, would have been inconsistent with the account; neither is observed.

\section{Detailed Finding B Numbers}
\label{app:findingB-table}

\begin{table*}[t]
\centering
\caption{Stage-2 rule $\times$ granularity under fixed DoReMi-L1 outer weights and HERMES-$256^3$ codebook, with both rules in their quota-preserving forms (Stage-1 L1 weights preserved exactly). At $L_{12}$ (median pool 2{,}271 docs), quality top-30\% outperforms coverage by $\Delta=+0.0253$ (Finding A). At $L_{123}$ (median pool 429 docs), the L12 advantage collapses: quality top-30\% and per-L1 $L_{123}$ local random coverage are numerically tied ($\Delta=+0.0002$, too small to interpret under single-seed evaluation), consistent with the candidate-competition mechanism (Section~\ref{sec:mechanism}). The global $L_{123}$ max-entropy row ($0.4061$) is a quota-flattening side-reference (Appendix~\ref{app:leaderboard}), not a quota-preserving Stage-2 contrast, and is not included here. All quality rows use the corrected FineWeb-Edu reader; single 1B/25B training seed per cell.}
\label{tab:findingB}
\small
\begin{tabularx}{\textwidth}{lXlXr}
\toprule
Granularity & Stage-2 rule & Mass dist. & Per-doc eligibility & Avg. \\
\midrule
$L_{12}$  & max-entropy (per-L1)   & equal sub-bucket   & all docs              & 0.3969 \\
$L_{12}$  & quality top-30\%       & size-proportional  & top-30\% by quality   & \textbf{0.4222} \\
$L_{123}$ & per-L1 local random    & equal sub-bucket   & all docs              & 0.3986 \\
$L_{123}$ & quality top-30\%       & size-proportional  & top-30\% by quality   & 0.3988 \\
\bottomrule
\end{tabularx}
\end{table*}

\section{Outer-Weight Characterisation at $L_1$}
\label{app:outerweight}

We report the paired $L_1$ comparison between our two $O(1)$-proxy outer-weight families (\emph{Uniform} and \emph{DoReMi}; Section~\ref{sec:sampling}) as a sanity audit, not as a structural finding. The paper's main granularity claim (Finding~A, Section~\ref{sec:findingA}) is conditioned on \emph{fixed} DoReMi outer weights; the comparison below characterises the cost of that choice at $L_1$, where we have the matched runs. We do not project these results to $L_{123}$, where the Uniform counterpart has not been run (see Section~\ref{sec:topline} and Limitations).

Four grouping families have been paired at L1 + L1 max-entropy, varying only whether the 256 L1 weights come from Uniform or DoReMi.

\begin{table*}[t]
\centering
\caption{Paired Uniform vs.\ DoReMi at $L_1$ + L1 max-entropy across four grouping families. We report the comparison descriptively; the sign at finer granularities is not directly testable from this table.}
\label{tab:outerweight}
\small
\begin{tabularx}{\textwidth}{Xrrrr}
\toprule
Grouping family & Uniform (Avg / z) & DoReMi (Avg / z) & $\Delta$ Avg & $\Delta$ z \\
\midrule
HERMES-$256^3$ ($c_1$, ours) & 0.4155 / $+1.048$ & 0.4028 / $+0.048$ & $-0.0127$ & $-1.000$ \\
KMeans-256 & 0.4153 / $+1.059$ & 0.3889 / $-1.147$ & $-0.0264$ & $-2.206$ \\
WebOrganizer-Format & 0.4134 / $+0.949$ & 0.4029 / $+0.128$ & $-0.0105$ & $-0.821$ \\
WebOrganizer-Topic & 0.4079 / $+0.469$ & 0.4082 / $+0.508$ & $+0.0003$ & $+0.039$ \\
\bottomrule
\end{tabularx}
\end{table*}

In three of four paired contrasts, replacing Uniform with DoReMi at $L_1$ reduces Avg.\ by $0.010$--$0.026$ accuracy points (z swing $-0.82$ to $-2.21$); WebOrganizer-Topic is numerically tied. We do not read this as a structural claim about DoReMi-style optimisers in general: this comparison is on a different axis from granularity, and the matched counterparts for $L_{12}$ quality top-30\% (the rank-1 row) have not been run.

\paragraph{Why DoReMi might underperform Uniform at $L_1$.} One natural reading is that group-DRO worst-group loss is misaligned with capability Avg.: DoReMi up-weights groups whose proxy-model loss is high, which at this scale tend to be format-heavy clusters (OCR fragments, code templates) that contribute little to the 16-task capability suite; geometric cluster groupings like KMeans amplify the misalignment (the largest $\Delta$ in the table). Whether this underperformance reverses at finer granularities or under different Stage-2 rules is open.

\section{Label-System Relationship}
\label{app:label-relationship}

We audit how HERMES labels relate to three reference label families on the intersection of documents annotated by all four systems ($n=259{,}255$): WebOrganizer Topic, WebOrganizer Format, and Topic-over-Source \citep{topicoversource}. Each reference family has $24$ classes, so the maximum reference entropy is $\log_2 24\approx 4.58$ bits. For each $(\text{HERMES prefix},\,\text{reference label})$ pair we report arithmetic-mean normalised mutual information (NMI), the median per-bucket purity (max-class fraction over the reference label, computed per HERMES bucket and aggregated by median across HERMES buckets), and conditional entropy $H(\text{reference}\mid\text{HERMES})$ in bits.

\begin{table}[t]
\centering
\caption{HERMES label system vs.\ three reference families on a $259{,}255$-document shared intersection. Each reference family has $24$ classes. Purity is the median across HERMES buckets of the max-class fraction over the reference label; $H(\cdot\mid\cdot)$ is in bits, against a $4.58$-bit reference-label prior. The shared intersection populates $52{,}536$ $L_{12}$ and $57{,}512$ $L_{123}$ HERMES buckets, fewer than the full-corpus totals (Table~\ref{tab:sampling}), because of the smaller document sample.}
\label{tab:label-relationship}
\footnotesize
\setlength{\tabcolsep}{4pt}
\begin{tabular}{@{}l l r r r@{}}
\toprule
HERMES & Reference & NMI & Purity & $H(\text{ref}|\text{H{\small ERMES}})$ \\
\midrule
\multirow{3}{*}{$L_1$}    & WO Topic  & 0.400 & 0.63 & 1.94 \\
                          & WO Format & 0.128 & 0.28 & 3.10 \\
                          & ToS       & 0.334 & 0.53 & 2.29 \\
\midrule
\multirow{3}{*}{$L_{12}$} & WO Topic  & 0.358 & 0.83 & 0.90 \\
                          & WO Format & 0.235 & 0.50 & 1.62 \\
                          & ToS       & 0.329 & 0.71 & 1.14 \\
\midrule
\multirow{3}{*}{$L_{123}$}& WO Topic  & 0.360 & 0.89 & 0.87 \\
                          & WO Format & 0.240 & 0.53 & 1.57 \\
                          & ToS       & 0.331 & 0.75 & 1.09 \\
\bottomrule
\end{tabular}
\end{table}

\paragraph{HERMES is not a clone of any reference family.} The highest NMI in Table~\ref{tab:label-relationship} is $0.40$ (HERMES $L_1$ vs.\ WO Topic), which is moderate, not identity. NMI with WO Format is the lowest ($0.13$ at $L_1$, $0.24$ at $L_{12}/L_{123}$); ToS sits in between ($0.33$ across all granularities). HERMES therefore captures a topical-leaning signal that overlaps but is not co-extensive with any of the three reference families.

\paragraph{Finer granularity sharpens topical alignment.} Median per-bucket purity climbs monotonically with HERMES granularity against every reference family: WO Topic $0.63\to 0.83\to 0.89$, ToS $0.53\to 0.71\to 0.75$, WO Format $0.28\to 0.50\to 0.53$. Conditional entropy of WO Topic given HERMES drops from $1.94$ bits at $L_1$ to $0.87$ bits at $L_{123}$, an $80\%$ reduction relative to the $4.58$-bit prior. At $L_{12}/L_{123}$ a HERMES bucket therefore behaves as a refined topical cluster: it does not name a single human-readable topic, but the documents inside it share one to a substantially greater extent than at $L_1$.

\paragraph{The format axis is comparatively under-resolved.} Even at $L_{123}$, median purity under WO Format is only $0.53$. HERMES's geometry-derived hierarchy captures topical content better than document format, consistent with the input embedding being a sentence-level semantic encoder rather than a layout/format model. A format-aware encoder is a natural future direction.

\paragraph{What this licenses.} HERMES at $L_{12}$ and $L_{123}$ can be read as a higher-resolution topical labeling, with the resolution exposed as a prefix length rather than rebuilt as a new clustering or taxonomy. Importantly, the moderate NMI with each of WO Topic, WO Format, and ToS at $L_1$ shows that HERMES does not duplicate any one reference taxonomy; the hierarchy adds resolution rather than re-deriving a fixed label system.

\section{Qualitative Label Inspection}
\label{app:label-inspection}

This appendix complements the quantitative label-relationship analysis (Appendix~\ref{app:label-relationship}) and the sparsity audit (Appendix~\ref{app:sampling}) with qualitative evidence that HERMES codes carry interpretable semantic content at every prefix level. All n-gram signatures below come from a uniformly spaced sample of $4{,}800{,}000$ documents ($160$ aligned text/annotation shards $\times$ $30{,}000$ rows per shard); full-corpus document counts come from the full $227$M-document annotation audit. For each L1 bucket, distinctive unigrams and bigrams were computed by ranking corpus-internal frequency against the corpus background. All $256$ L1 buckets are populated in the sample (per-bucket sample size $3{,}481$ to $52{,}912$, median $18{,}160$).

\subsection{Representative L1 buckets}
\label{app:label-inspection-l1}

\begin{table*}[t]
\centering
\caption{Representative HERMES L1 bucket signatures from a $4.8$M-document sample. The full $256$-row table is released in the supplementary material. The examples show that the learned labels recover coherent topical and stylistic regions without using hand-written taxonomies; per-bucket interpretable labels are author-assigned summaries of the n-gram signatures, not learned outputs.}
\label{tab:label-inspection-l1}
\footnotesize
\setlength{\tabcolsep}{5pt}
\renewcommand{\arraystretch}{1.15}
\begin{tabularx}{\textwidth}{@{}r l r r >{\raggedright\arraybackslash}X@{}}
\toprule
L1 cell & Interpretable label & Full-corpus docs & Sample docs & Distinctive n-grams (top 6) \\
\midrule
179 & Books, publishing, fiction         & 2{,}525{,}694 & 52{,}912 & contemporary romance, cover reveal, netgalley, urban fantasy, debut novel, hardcover paperback \\
24  & Music releases and reviews         & 2{,}102{,}386 & 45{,}359 & album review, second album, progressive rock, released album, title track, debut single \\
218 & Visual art and exhibitions         & 1{,}797{,}386 & 37{,}901 & solo exhibition, museum contemporary, artist statement, painting sculpture, art practice, art fair \\
11  & Software and developer tooling     & 1{,}725{,}904 & 28{,}991 & version control, sqlite, dbforge, configuration file, unit tests, jdbc \\
6   & Recipes and cooking                & 1{,}280{,}176 & 26{,}562 & marinade, cook minutes, finely chopped, pepper taste, cloves garlic, saute \\
4   & Biomedicine, molecular biology     & 1{,}255{,}632 & 25{,}967 & gene expression, assays, crispr, gene therapy, kinase, neuronal \\
8   & Video games and RPGs               & 1{,}050{,}568 & 21{,}781 & azeroth, guild wars, npcs, edh, pve, clash royale \\
116 & Astronomy and space science        &   890{,}156 & 17{,}993 & astronomers, nasa's, hubble, space agency, black holes, cassini \\
10  & Macroeconomics and markets         &   885{,}835 & 19{,}210 & gdp growth, bernanke, bull market, yellen, yield curve, fomc \\
97  & Wine and winemaking                &   617{,}996 & 12{,}830 & winemaker, winemaking, cabernet sauvignon, tannins, sauvignon blanc, riesling \\
76  & Eye care and ophthalmology         &   431{,}300 &  9{,}033 & lasik, cornea, optometry, macular, contact lens, cataracts \\
254 & North Korea / geopolitics          &   157{,}314 &  3{,}481 & pyongyang, kim jong, korean peninsula, jong-un, kim jong-un, korean leader \\
\bottomrule
\end{tabularx}
\end{table*}

\subsection{Prefix hierarchy examples}
\label{app:label-inspection-hierarchy}

To check that prefix length acts as a semantic granularity control rather than a numeric handle, we selected four high-population L1 parents and inspected their largest L12 children, then drilled into selected L123 descendants. Parent and child summaries are author-assigned from the n-gram signatures.

\begin{table*}[t]
\centering
\caption{Representative HERMES prefix hierarchy. For each of four high-population L1 parents we show its five largest L12 children. ``Active L12'' counts L12 sub-buckets that received any documents under the parent; child summaries are author-assigned from n-gram signatures on the $4.8$M-document sample.}
\label{tab:label-inspection-hierarchy}
\footnotesize
\setlength{\tabcolsep}{5pt}
\renewcommand{\arraystretch}{1.2}
\begin{tabularx}{\textwidth}{@{}l r r l >{\raggedright\arraybackslash}X@{}}
\toprule
L1 parent & Full docs & Active L12 & L12 child & Summary \\
\midrule
\multirow{5}{*}{179: Books / publishing}  & \multirow{5}{*}{2{,}525{,}694} & \multirow{5}{*}{256}
  & 179\_97  & writing/novel advice \\
&&& 179\_176 & children's books \\
&&& 179\_255 & romance \\
&&& 179\_148 & sci-fi/fantasy \\
&&& 179\_133 & classic/animal literature \\
\midrule
\multirow{5}{*}{24: Music}  & \multirow{5}{*}{2{,}102{,}386} & \multirow{5}{*}{256}
  & 24\_18  & classic rock \\
&&& 24\_65  & music videos \\
&&& 24\_150 & MP3/download pages \\
&&& 24\_12  & hip-hop / pop artists \\
&&& 24\_183 & Christian / gospel music \\
\midrule
\multirow{5}{*}{218: Visual art}  & \multirow{5}{*}{1{,}797{,}386} & \multirow{5}{*}{255}
  & 218\_9   & general art discourse \\
&&& 218\_199 & art education \\
&&& 218\_18  & modern artists \\
&&& 218\_243 & Renaissance / Italian art \\
&&& 218\_120 & Islamic / Middle Eastern art \\
\midrule
\multirow{5}{*}{11: Software / dev tools}  & \multirow{5}{*}{1{,}725{,}904} & \multirow{5}{*}{256}
  & 11\_235 & Azure / ASP.NET / API docs \\
&&& 11\_150 & file/PDF tooling \\
&&& 11\_226 & data/table tooling \\
&&& 11\_123 & algebra/math software \\
&&& 11\_142 & unit and functional testing \\
\bottomrule
\end{tabularx}
\end{table*}

\paragraph{L123 drill-down.} A selected L123 inspection is consistent with the sparsity audit (Appendix~\ref{app:sampling}): where the third RVQ stage is populated, it surfaces local refinements rather than a dense third-level tree. For example, the hip-hop / pop-artist sub-cluster $24\_12$ splits into $24\_12\_250$ (popcaan, drizzy, drake, ras cal) and $24\_12\_95$ (cudi, chioma, quavo, archuleta), and the art-education sub-cluster $218\_199$ splits into $218\_199\_127$ (art students, art academy, teaching art) and $218\_199\_16$ (art students, art teachers, scholastic art).

\paragraph{Scope and caveat.} These examples are qualitative and selected for readability; they support the interpretation of HERMES codes as a prefix-readable hierarchical label substrate, and are consistent with the quantitative purity climb (Appendix~\ref{app:label-relationship}) and the parent-child refinement statistics (Appendix~\ref{app:sampling}). They are not a downstream-performance result, and not every bucket is equally clean: e.g., L1 buckets sitting near format/quality boundaries (OCR fragments, navigation pages) are not represented in this selection but appear in the full $256$-row table.

\end{document}